\documentclass[IROS]{RLIpaper}

\usepackage{geometric_algebra}

\begin{document}

\title{\textit{gafro}: Geometric Algebra for Robotics}
\author{\IEEEauthorblockN{Tobias Löw\IEEEauthorrefmark{1}\IEEEauthorrefmark{2}, Philip Abbet\IEEEauthorrefmark{1} and Sylvain Calinon\IEEEauthorrefmark{1}\IEEEauthorrefmark{2}}
\IEEEauthorblockA{\IEEEauthorrefmark{1}Idiap Research Institute, Martigny, Switzerland}
\IEEEauthorblockA{\IEEEauthorrefmark{2}EPFL, Lausanne, Switzerland}
}

\maketitle

\begin{abstract}
Geometry is a fundamental part of robotics and there have been various frameworks of representation over the years. Recently, geometric algebra has gained attention for its property of unifying many of those previous ideas into one algebra. While there are already efficient open-source implementations of geometric algebra available, none of them is targeted at robotics applications. We want to address this shortcoming with our library \textit{gafro}. This article presents an overview of the implementation details as well as a tutorial of \textit{gafro}, an efficient c++ library targeting robotics applications using geometric algebra. The library focuses on using conformal geometric algebra. Hence, various geometric primitives are available for computation as well as rigid body transformations. The modeling of robotic systems is also an important aspect of the library. It implements various algorithms for calculating the kinematics and dynamics of such systems as well as objectives for optimisation problems. The software stack is completed by python bindings in \textit{pygafro} and a ROS interface in \textit{gafro\_ros}.
\end{abstract}


\section{INTRODUCTION}
\label{sec:introduction}
    \addgrant{This work was supported by the State Secretariat for Education, Research and Innovation in Switzerland for participation in the European Commission’s Horizon Europe Program through the INTELLIMAN project (https://intelliman-project.eu/, HORIZON-CL4-Digital-Emerging Grant 101070136) and the SESTOSENSO project (http://sestosenso.eu/, HORIZON-CL4-Digital-Emerging Grant 101070310).}

    Geometric algebra can be considered a high-level mathematical language for geometric reasoning. As such it is very well suited for general problems in robotics.

    Although geometric algebra has great potential for modeling, learning and control in robotics, it has not been widely adopted in robotics research. One reason for this is the lack of easy-to-use libraries for robotics applications, while at the same time tools based on matrix algebra are very mature and readily available. We aim to change that by providing a ready-to-use geometric algebra library for robotics that can be used with the most popular programming frameworks, namely c++, python and ROS.

    For robot modeling and control it is necessary to compute the kinematics and dynamics of robots. These algorithms are well studied and have been implemented in various software frameworks. In our \textit{gafro} library we provide an implementation of the geometric algebra variant of these algorithms. It is important to point out that geometric algebra can be used to compute these important quantities that are classically computed using matrix algebra, while also offering a richer toolset, i.e. it also includes tools for geometric reasoning that matrix algebra does not have. 

    In this article we want to explain the implementation details of our geometric algebra library \textit{gafro} and give a tutorial on how to use it for common robotics problems such as inverse kinematics and optimal control. Our aim is to make geometric algebra more accessible for robotics research by providing this ready-to-use library. This should result in a wider adoption and facilitate the research on using this powerful framework for robotics. 

    This article is organized as follows: in Section \ref{sec:programming_interfaces} we give an overview of the available programming interfaces, in Section \ref{sec:conformal_geometric_algebra_implementation} we explain the implementation details of the algebra, in Section \ref{sec:robot_modeling} we show how to model robots using the library, in Section \ref{sec:comparison_to_other_libraries} we compare \textit{gafro} to other GA and robot modeling libraries and finally in Section \ref{sec:applications} we demonstrate various applications and give a tutorial on how to use the library. The documentation and the links to all repositories can be found on our website \texttt{\url{https://geometric-algebra.tobiloew.ch/gafro}}.


\section{PROGRAMMING INTERFACES}
\label{sec:programming_interfaces}
    In this section we explain the programming interfaces that \textit{gafro} offers. The main library is written in c++ for which we provide python bindings called \textit{pygafro} as well as the ROS package \textit{gafro\_ros}. All mentioned repositories can be found at \texttt{\url{https://gitlab.com/gafro}}.

    
    \subsection{c++ Library}
    \label{sub:library}
        The core implementation of \textit{gafro} is done in c++20 and relies heavily on templates. Apart from the standard library the only mandatory dependency is the \textit{Eigen}\footnote{\url{https://eigen.tuxfamily.org}} library. We use the the \textit{yaml-cpp}\footnote{\url{https://github.com/jbeder/yaml-cpp}} library as an optional dependency for loading of robot descriptions. Several robot descriptions are already available and can be used via the respective classes. 
    
    \subsection{ROS Package}
    \label{sub:ros_package}
        The Robot Operating System (ROS) is the de facto standard for building robot applications. Hence, we provide a ROS package called \textit{gafro\_ros} to use our library within the ROS framework. It allows passing geometric primitives between nodes via messages. It also enables the visualization of these geometric primitives in Rviz via custom plugins. Furthermore, since most robot descriptions are available in the Universal Robot Description Format (URDF), this package supports loading this file format.

    
    \subsection{Python Bindings}
    \label{sub:python_bindings}
        Since the python programming language is a popular tool for rapid prototyping and is generally more accessible than the c++ language, we are providing language bindings in python to \textit{gafro} using the \textit{pybind11}\footnote{\url{https://github.com/pybind/pybind11}} library. 


\section{IMPLEMENTATION OF CONFORMAL GEOMETRIC ALGEBRA}
\label{sec:conformal_geometric_algebra_implementation}
    In this section we will explain in detail our implementation of conformal geometric algebra (CGA). The aspects that are highlighted are the implementation of a general multivector and the expressions that are acting on it. The library covers several points that were proposed in \cite{bengerMassiveGeometricAlgebra2017} as a wishlist of geometric algebra implementations. We have designed the library in an object-oriented way, so the classes also reflect the mathematical inheritance relationships. Furthermore all classes, i.e. all specialized multivectors are instantiated as different types, which allows them to be distinguished at compile-time for type-safety and have persistent storage. These specialized classes, also enable the computation with partial multivectors, i.e. the library exploits the fact that the most commonly mulivectors are sparse and only use certain subspaces of the algebra. We address numerical imprecisions by ensuring that only elements of the resulting multivectors of expressions are evaluated that are known to be non-zero. We handle the type explosion of binary operators by automatically evaluation partial expressions when the full expressions get too complex. 
    
    
    \subsection{General Multivector}
    \label{sub:general_multivector}
\begin{figure}[!ht]
            \centering
            \includegraphics[width=0.9\linewidth]{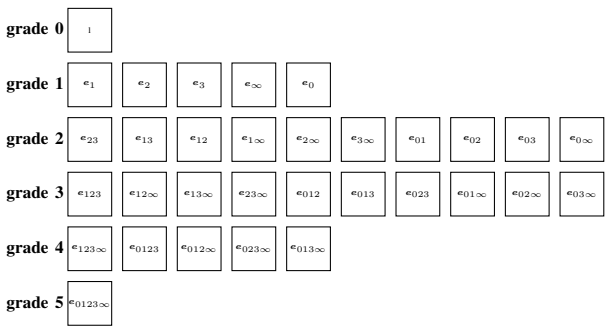}
            \caption{Structure of conformal geometric algebra with the 32 basis blades, divided into the different grades. Grade 0 and 5 are the scalar and pseudo-scalar, respectively. Grades 1 to 4 are called bi-, tri- and quadvectors.}
            \label{fig:structure_of_conformal_geometric_algebra}
        \end{figure}

        The core element of computation in geometric algebra is the multivector. Hence, it is very important to think about the design choices when implementing its structure, as this will determine the memory usage and computational performance. The general structure of a multivector in CGA can be seen in Figure \ref{fig:structure_of_conformal_geometric_algebra}. It is composed of 32 basis blades, divided into grades zero to five. A general multivector would therefore be quite heavy in terms of memory and computation. However, 

        \begin{figure}[!ht]
            \centering
            \begin{minipage}{0.55\linewidth}
                \centering
                \includegraphics[width=0.75\linewidth]{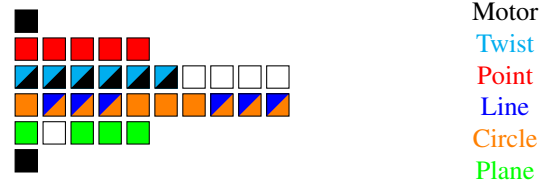}
            \end{minipage}
            \hfill
            \begin{minipage}{0.39\linewidth} 
                \centering
                \textcolor{black}{Motor}
                \\ \textcolor{cyan}{Twist}
                \\ \textcolor{red}{Point}
                \\ \textcolor{blue}{Line}
                \\ \textcolor{orange}{Circle}
                \\ \textcolor{green}{Plane}
            \end{minipage}
            \caption{Non-zero elements of various geometric primitives in their primal representations in conformal geometric algebra. Boxes represent basis blades and colored boxes represent the non-zero blades of the geometric primitive with the matching color. It can be seen that of the 32 basis blades composing multivectors only a sparse number is used for the representations. Note that geometric primitives are single-grade objects, while transformations are mixed-grade.}
            \label{fig:structure_of_various_geometric_primitives_in_conformal_geometric_algebra_}
        \end{figure}

        An important structural aspects of CGA that facilitate the design process here are the sparsity of its representations and the fact that the structure of multivector expressions is known at compile-time. Both of these properties mean that we can implement the data vector of a multivector by only storing its known non-zero elements. This is achieved by using a template that takes list of blade indices as input:

\begin{lstlisting}
template <class T, int... blades>
class Multivector
{
    public:
        constexpr static int size = sizeof...(blades);
        {...}
    private:
        Eigen::Matrix<T, size, 1> data_;
};
\end{lstlisting}

        \noindent The list of indices is then stored internally as a bitset that facilitates the comparison of the subspaces of two mulivectors. A bitset is simply a list of 32 bits, that are either 0 or 1, depending on whether the corresponding blade is present in the multivector or not.

        The memory that is allocated corresponds to the number of blade indices that is given to the template. It uses an \texttt{Eigen::Matrix} to store the data, that is exposed via an accessor function called \texttt{vector()}. This makes it possible to directly use the parameter vector of any multivector, which is useful for e.g. optimisation solvers. 

        The underlying data type \texttt{T} is a template argument, which makes it possible to either use e.g. \texttt{float} or \texttt{double}, depending on the system architecture. Furthermore, it allows the usage of general purpose automatic differentiation libraries such as \textit{autodiff}\footnote{\url{https://autodiff.github.io/}}. This helps when formulizing optimisation problems in geometric algebra using \textit{gafro} since it facilitates the coding of complex objective functions and thus accelerates prototyping.
            
        \begin{table}[!ht]
            \centering
            \caption{Unary expressions that are implemented as member functions of the \texttt{Multivector} class.}
            \begin{tabular}{lll}
                \toprule
                    \textbf{member function}
                    & \textbf{unary expression}
                    & \textbf{mathematical symbol}
                    \\
                \midrule
                    \texttt{reverse}
                    & \texttt{Reverse}
                    & $\reverse{X}$
                    \\
                    \texttt{inverse}
                    & \texttt{Inverse}
                    & $X^{-1}$
                    \\
                    \texttt{dual}
                    & \texttt{Dual}
                    & $X^*$
                    \\
                \bottomrule
            \end{tabular}
            \label{tab:unary_expressions_that_are_implemented_as_member_functions_of_the_multivector_class_}
        \end{table}

        The \texttt{Multivector} class and all its derivates (including the expressions) have a method called \texttt{get} that is templated on the blade index and protected by the concepts library of c++20. 

    
    \subsection{Algebraic Computations using Expression Templates}
    \label{sub:algebraic_computations_using_expression_templates}
        When implementing geometric algebra, there are some core binary expressions that need to be available for general algebraic computations. These operators are listed in Table \ref{tab:binary_operators_}. 

        \begin{table}[!ht]
            \centering
            \caption{Binary operators.}
            \begin{tabular}{lcc}
                \toprule
                    & \textbf{operator}
                    & \textbf{symbol}
                    \\
                \toprule
                    addition
                    & +
                    & $+$
                    \\
                    substraction
                    & -
                    & $-$
                    \\
                    outer product
                    & $\hat{}$
                    & $\outer$ 
                    \\
                    inner product
                    & $|$
                    & $\inner$ 
                    \\
                    geometric product
                    & $*$
                    & 
                    \\
                \bottomrule
            \end{tabular}
            \label{tab:binary_operators_}
        \end{table}

        All expressions are implemented as expression templates. The expression templates determine the result type of the expression at compile time. This is achieved via accompanying evaluation classes that are hidden in the \texttt{detail} namespace. The challenge in the implementation here is the fact that the resulting multivectors only rarely have the same blades as the input operands. Note that the expressions are evaluated in a lazy fashion, which means that the blades are evaluated on demand. This makes it possible to e.g. only evaluate a single blade of the resulting multivector, depending on the requirements.

\begin{lstlisting}
template <class Derived, class Result>
class Expression
{
    public:
        template <int blade>
        requires(Result::has(blade))
        typename Result::Vtype get() const
        {
            return static_cast<const Derived &>(*this).template get<blade>();
        }
};
\end{lstlisting}

\begin{figure}[!ht]
            \centering
            \begin{subfigure}[]{\linewidth}
                \centering
                \includegraphics[width=\linewidth]{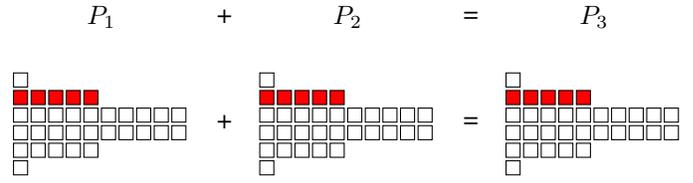}
                \caption{The summation of two mulivectors with the same blades results in another multivector with the same blades.}
            \end{subfigure}

            \begin{subfigure}[]{\linewidth}
                \centering
                \includegraphics[width=\linewidth]{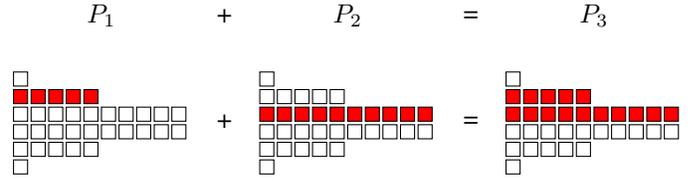}
                \caption{The summation of two mulivectors with the different blades results in a multivector with the blades of both input mulivectors.}
            \end{subfigure}
            \caption{Summation}
            \label{fig:summation_expression}
        \end{figure}

        A first example of this can be seen in the \texttt{Sum} expression in Figure \ref{fig:summation_expression}. The corresponding type evaluation class constructs the type of the resulting multivector at compile-time. In the case of a summation this amounts to a simple bitwise OR operation comparing the bitsets of the input multivectors.
        
\begin{figure}[!ht]
            \centering 
            \begin{subfigure}[]{\linewidth}
                \centering
                \includegraphics[width=\linewidth]{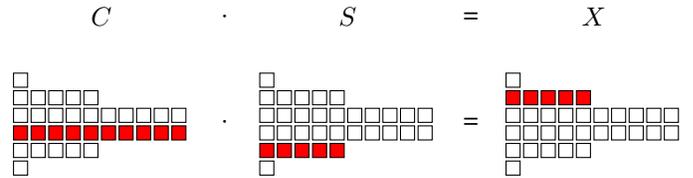}
                \caption{The inner product is a grade lowering operation, i.e. the resulting multivector will be of lower grade than the inputs. The example shows that the inner product of a circle $C$ with a sphere $S$ results in a point $P$.}
            \end{subfigure}

            \begin{subfigure}[]{\linewidth}
                \centering
                \includegraphics[width=\linewidth]{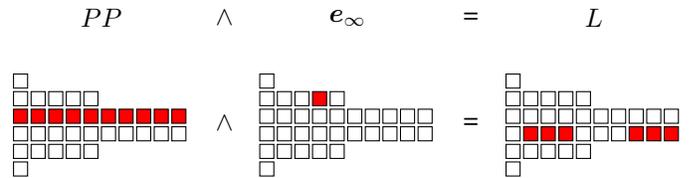}
                \caption{The outer product is a grade raising operation, i.e. the resulting multivector will be of higher grade than the inputs. The example shows that the outer product of a pointpair $PP$ and $\gae{\infty}$ results in a line $L$}
            \end{subfigure}
            \caption{The resulting multivector of the inner and outer product operations has a different grade than the inputs.}
            \label{fig:inner_outer_product_expression}
        \end{figure}

        The inner and outer products work essentially in the same way and thus the corresponding expression both inherit from a base \texttt{Product} class, i.e. \texttt{InnerProduct} and \texttt{OuterProduct}. The \texttt{Product} class takes a class structure implementing the corresponding Cayley table as template argument. This Cayley table defines the resulting blades of a blade by blade multiplication under the inner and outer product, respectively. Thus, in the case of CGA, it defines 1024 operations. In order to determine the type of the resulting multivector, we employ fold expressions that allow us to iterate over the blades of both input mulivectors at compile-time. In this loop, we obtain the resulting blade per pair of blades using the respective Cayley table and then assemble them into the resulting multivector again using OR operations. Figure \ref{fig:inner_outer_product_expression} shows an example for each the inner and the outer product. 

        The geometric product class \texttt{GeometricProduct} also inherits from the base \texttt{Product} class and comes with its own Cayley table. So implementation wise it is the same as the inner and outer products. The main difference is that two blades can result from a blade multiplication, which causes the resulting multivector to potentially have both a lower and a higher grade than the inputs, as can be seen in Figure \ref{fig:geometric_product}. Furthermore, we have products that are based on the geometric product such as the sandwich product, which is also treated as a binary expression and shown in Figure \ref{fig:sandwich_product}.

        \begin{figure}[!ht]
            \centering
            \begin{subfigure}[]{\linewidth}
                \centering
                \includegraphics[width=\linewidth]{images/geometric_product_binary}
                \caption{Using the geometric product results in both blades of lower and higher grade.}
                \label{fig:geometric_product}
            \end{subfigure}

            \begin{subfigure}[]{\linewidth}
                \centering
                \includegraphics[width=\linewidth]{images/geometric_product_motor}
                \caption{The sandwich product is a grade-preserving operation. Numerical issues might lead to residuals in other blades, which we avoid by simply not evaluating them in the expressions.}
                \label{fig:sandwich_product}
            \end{subfigure}
            \caption{The geometric product is a combination of the inner and outer product.}
        \end{figure}


    
    \subsection{Geometric Primitives}
    \label{sub:geometric_primitives}
        Since we know the subspaces of all the geometric primitives, we chose to implement them in an object-oriented way by inheriting from the base \texttt{Multivector} class. Hence, the available classes are \texttt{Vector}, \texttt{DirectionVector}, \texttt{TangentVector}, \texttt{Point}, \texttt{PointPair}, \texttt{Line}, \texttt{Circle}, \texttt{Plane} and \texttt{Sphere}. Their corresponding subspaces within the geometric algebra can be seen in Figure \ref{fig:structure_of_various_geometric_primitives_in_conformal_geometric_algebra_}. Having the geometric primitives as explicit classes allows the implementation of commonly used equations as members functions, which facilitates the usage. For example the constructors of the geometric primitive classes implement the various ways they can be defined. The explicit classes are meant to facilitate the use and construction, but of course, using computation with base multivectors is also possible. This preserves the property of covariant computation within the algebra.

    
    \subsection{Rigid Body Transformations}
    \label{sub:rigid_body_transformations}
        The rigid body transformations that are currently available are implemented in the classes \texttt{Rotor}, \texttt{Translator}, \texttt{Motor} and \texttt{Dilator}. They all inherit from the base class \texttt{Versor}. Since all three classes are exponential mappings of bivectors they are accompanied by the expressions \texttt{*Logarithm} and \texttt{*Exponential}, respectively. The main method of the rigid body transformations is \texttt{apply}, which implements the sandwich product $X'=MX\reverse{M}$ and ensures type safety. While $X$ can technically be any multivector, the intended usage is with the geometric primitives that were presented in Section \ref{sub:geometric_primitives}. Hence, in this context, type safety means that $X'$ stays the same geometric primitive as $X$, e.g. a \texttt{Point} stays a \texttt{Point}. This ensures that the expression only evaluates blades that are part of the geometric primitive, which not only reduces the number of floating point operations, it also deals with numerical imprecisions in the computation that are known to occur in geometric algebra implementations. 


\section{ROBOT MODELING}
\label{sec:robot_modeling}
    The previous sections introduced the features related to the underlying geometric algebra implementation of \textit{gafro}. This section will now introduce the higher level features of the library related to robot modeling, that distinguish it from other geometric algebra libraries. 

    The main aspects of robot modeling are the computation of the kinematics and dynamics of robotic systems. While the forward kinematics and forward/inverse dynamics can be computed using efficient recursive algorithms, the inverse kinematics problem for redundant manipulators is an optimisation problem. Hence, the robot modeling section also covers the necessary tools for solving such optimisation problems.

    The base class for the modeling of robotic systems is called \texttt{System}. It implements the main functionality for the computation of the kinematics and dynamics of robotic systems and stores the joints and links. Those are organized in the classes \texttt{Joint}, \texttt{FixedJoint}, \texttt{RevoluteJoint}, \texttt{PrismaticJoint} and \texttt{Link}. A custom robot system can then be created using the functions \texttt{addJoint} and \texttt{addLink}. Note that those functions only add the links and joints to the system, the proper relationships between them need to be specified individually by setting the respective child/parent links/joints. Usually, robotic systems consist of one or more kinematic chains of interests, e.g. the chain from the base link to the end-effector frame for serial manipulators. This concept is implemented in the \texttt{KinematicChain} helper class and can be added to a \texttt{System} via the \texttt{addKinematicChain} function.

    The forward and inverse dynamics are implemented in the member functions \texttt{computeJointAccelerations} and \texttt{computeJointTorques}, respectively. These functions are using the conformal geometric algebra versions of the Recursive Newton-Euler Algorithm (RNEA) and Articulated Body Algorithm (ABA). 

    The \texttt{Manipulator} class that implements a serial manipulator with $N$ degrees of freedom is a specialization of the base \texttt{System} class. The parameter $N$ is given as a template argument, which makes it possible to use fixed-size \texttt{Eigen::Matrix} types for the implementation of e.g. the joint configuration vectors and Jacobian matrices. The \texttt{Manipulator} class implements functions that are most commonly of interest when dealing with such types of systems. More specifically, these are functions dealing with the end-effector forward kinematics and Jacobian matrices. Hence, these functions are
    \begin{itemize}
        \item \texttt{getEEMotor}
        \item \texttt{getEEAnalyticJacobian}
        \item \texttt{getEEGeometricJacobian}
        \item \texttt{getEEFrameJacobian}
    \end{itemize}
    These functions use the aforementioned \texttt{KinematicChain} class in order to define the end-effector kinematic chain and precompute certain related quantities. Hence, loading a \texttt{Manipulator} from a file addionally requires the name of the end-effector joint.
    
    The \textit{gafro} library includes several predefined serial manipulators that inherit from the \texttt{Manipulator} class, those are implemented in the following classes: 
    \begin{itemize}
        \item \texttt{FrankaEmikaRobot}
        \item \texttt{UR5}
    \end{itemize}
    These classes are loading the definition from yaml files and are only available if \textit{yaml-cpp} is installed on the system. The ROS package contains a program that converts URDF files to the yaml format that is being used by this library. 


\section{COMPARISON TO OTHER LIBRARIES}
\label{sec:comparison_to_other_libraries}
    There have been various works that published implementations of geometric algebra. These libraries all have in common that they are meant to be generic geometric algebra implementations focusing on the computational and mathematical aspects of the algebra itself. In contrast to that, our implementation is targeted specifically at robotics applications and thus not only implements the low-level algebraic computations but also features the computation of the kinematics and dynamics of serial manipulators as well as generic cost functions for optimal control.

    \begin{table}[!ht]
        \centering
        \caption{Comparison of different libraries.}
        \begin{subtable}[h]{\linewidth}
            \caption{Overview of other geometric algebra libraries.}
            \begin{tabularx}{\linewidth}{lcX}
                \toprule
                    \textit{Garamon}
                    & \cite{breuilsGaramonGeometricAlgebra2019}
                    & a generator of C++ libraries dedicated to geometric algebra
                    \\
                \midrule
                    \textit{GATL}
                    & \cite{fernandesExploringLazyEvaluation2021}
                    & C++ library for Euclidean, homogeneous/projective, Minkowski/spacetime, conformal, and arbitrary geometric algebras using template meta-programming
                    \\
                \midrule
                    \textit{Versor}
                    & \cite{colapinto2011versor}
                    & (fast) generic C++ library for geometric algebras
                    \\
                \midrule
                    \textit{GAL}
                    & \cite{jeremy_ong_2019}
                    & C++17 expression compiler and engine for computing with geometric algebra
                    \\
                \midrule
                    \textit{Gaigen}
                    & \cite{fontijneGaigenGeometricAlgebra2006}
                    & code generator for geometric algebra
                    \\
                \midrule
                    \textit{Gaalet}
                    & \cite{seyboldGaaletExpressionTemplate2010}
                    & C++ library for evaluation of geometric algebra expressions offering comfortable implementation and reasonable speed by using expression templates and metaprogramming techniques
                    \\
                \midrule
                    \textit{Gaalop}
                    & \cite{Hildenbrand2010}
                    & software to optimize geometric algebra files
                    \\
                \midrule
                    \textit{TbGAL}
                    & \cite{sousaTbGALTensorBasedLibrary2020}              
                    & C++/Python library for Euclidean, homogeneous/projective, Minkowski/spacetime, conformal, and arbitrary geometric algebras representing blades (and versors) in their decomposed state to scale to scale high dimensions
                    \\
                \bottomrule
            \end{tabularx}
        \end{subtable}

        \vspace{5mm}

        \begin{subtable}[h]{\linewidth}
                \caption{Overview of other libraries for robot modeling.}
                \begin{tabularx}{\linewidth}{lcX}
                    \toprule
                        \textit{DQ Robotics} 
                        & \cite{adornoDQRoboticsLibrary2021}
                        & library for robot modeling and control based on dual quaternion algebra
                        \\
                    \midrule
                        \textit{Pinocchio}
                        & \cite{carpentierPinocchioLibraryFast2019}
                        & state-of-the-art rigid body algorithms for poly-articulated systems
                        \\
                    \midrule
                        \textit{Raisim}
                        & \cite{hwangboPerContactIterationMethod2018}
                        & multi-body physics engine for robotics and AI
                        \\
                    \midrule
                        \textit{KDL}
                        & \cite{smitsKDLKinematicsDynamics}
                        & application independent framework for modeling and computation of kinematic chains
                        \\
                    \midrule
                        \textit{Mujoco}
                        & \cite{todorovMuJoCoPhysicsEngine2012}
                        & physics engine for model-based optimisation
                        \\
                    \midrule
                        \textit{RBDL}
                        & \cite{felisRBDLEfficientRigidbody2017}
                        & highly efficient code for both forward and inverse dynamics for kinematic chains and branched models
                        \\
                    \bottomrule
                \end{tabularx}
        \end{subtable}
        
        \label{tab:comparison_of_different_libraries}
    \end{table}

    
    \subsection{Algebraic Operations Benchmarks}
    \label{sub:benchmarks}
        In order to compare the performance of our \textit{gafro} library to other geometric algebra libraries we forked the \textit{ga\_benchmark}\footnote{\url{https://github.com/ga-developers/ga-benchmark}} repository in order to integrate \textit{gafro}. Our fork can be found at \texttt{\url{https://github.com/loewt/ga-benchmark}}.

        \begin{figure}[!ht]
            \begin{subfigure}[t]{\linewidth}
                \centering
                \includegraphics[width=\linewidth]{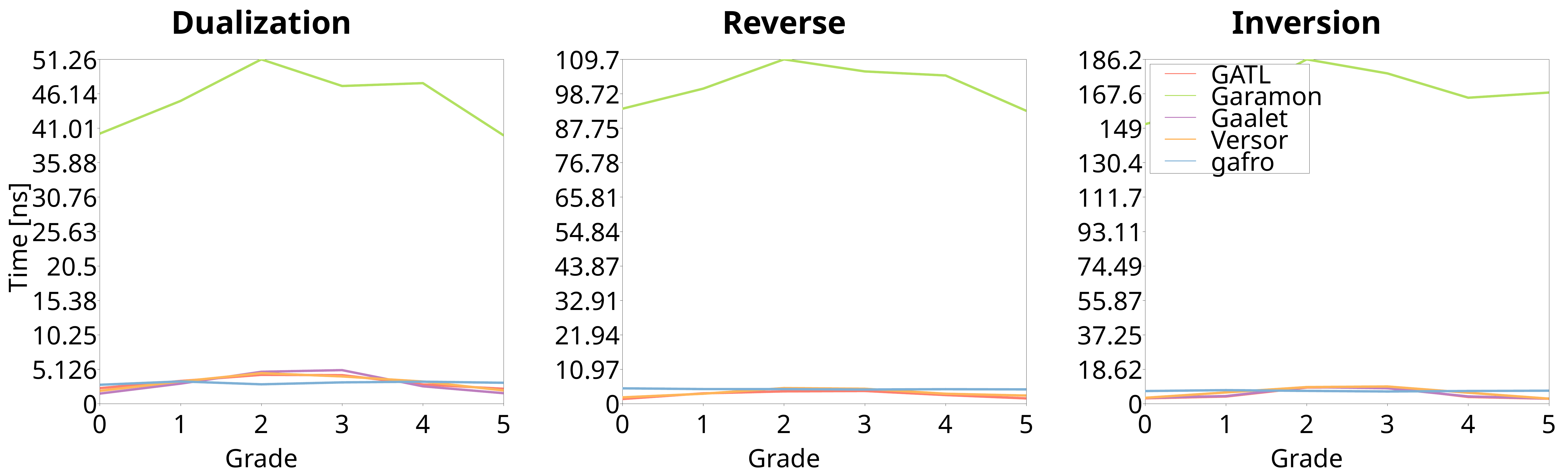}
                \caption{Benchmarks of unary algebraic operations.}
                \label{fig:benchmarks_of_unary_algebraic_operations_}
            \end{subfigure}
            
            \begin{subfigure}[t]{\linewidth}
                \centering
                \includegraphics[width=\linewidth]{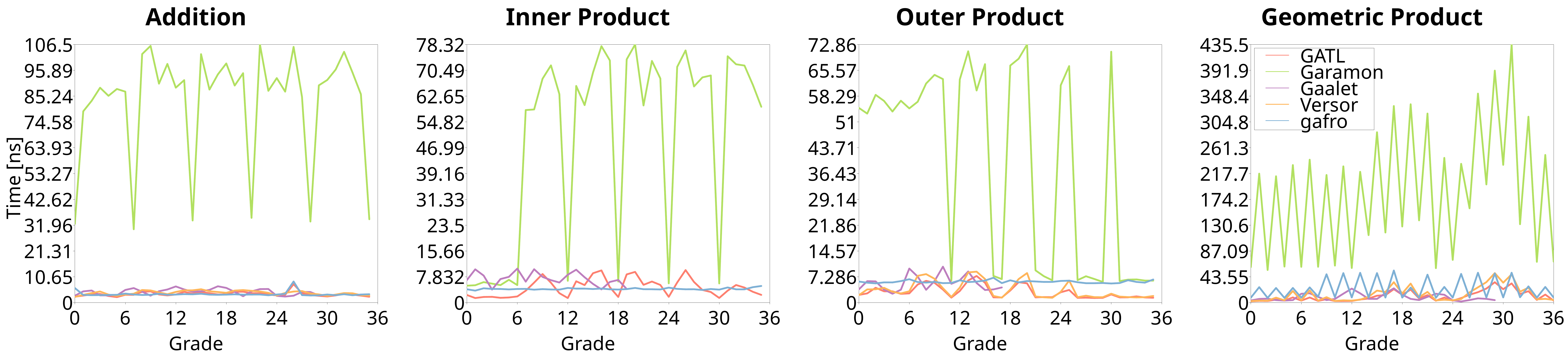}
                \caption{Benchmarks of binary algebraic operations.}
                \label{fig:benchmarks_of_binary_algebraic_operations_}
            \end{subfigure}
            \caption{Benchmarks of different geometric algebra libraries. All operations are computed using conformal geometric algebra.}
            \label{fig:benchmarks_of_geometric_algebra_libraries}
        \end{figure}

        We omitted \textit{TbGAL} from the plots of the benchmark results, since it is by far the slowest library. The benchmarks show that \textit{gafro} can compete in terms of performance with \textit{GATL} and \textit{Versor}, which were previously reported to be the fastest GA libraries. We believe, however, that the API of \textit{gafro} is a lot more approachable and easy to use. Furthermore, it additionally implements specialized algorithms for robot modeling and optimisation problems.

    
    \subsection{Robotics Algorithms Benchmarks}
    \label{sub:robotics_algorithms_benchmarks}
\begin{figure}[!ht]
            \centering
            \begin{subfigure}[t]{0.475\linewidth}
                \includegraphics[width=\linewidth]{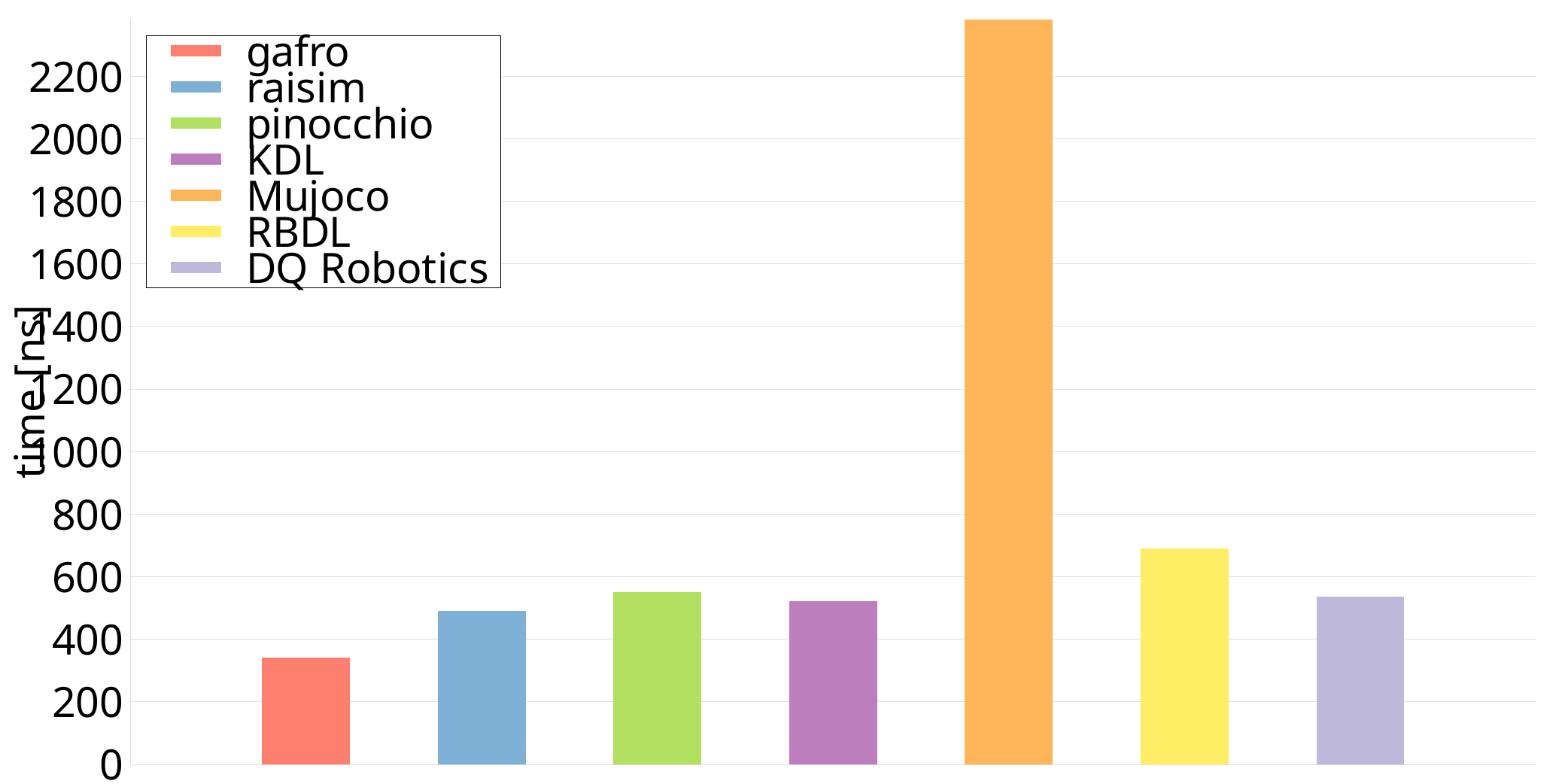}
                \caption{Forward Kinematics.}
            \end{subfigure}
            \hfill 
            \begin{subfigure}[t]{0.475\linewidth}
                \includegraphics[width=\linewidth]{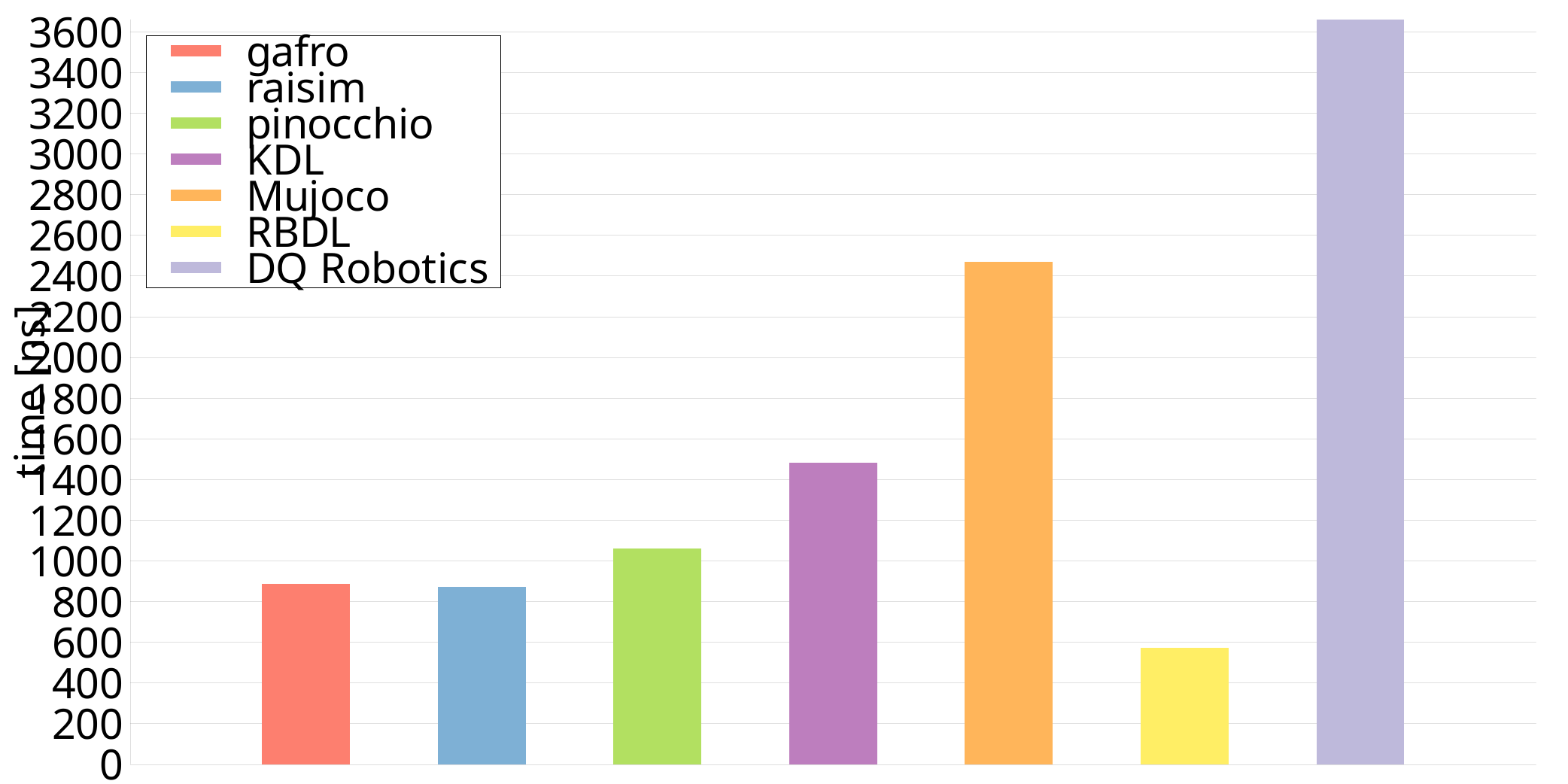}
                \caption{Jacobian.}
            \end{subfigure}
            \begin{subfigure}[t]{0.475\linewidth}
                \includegraphics[width=\linewidth]{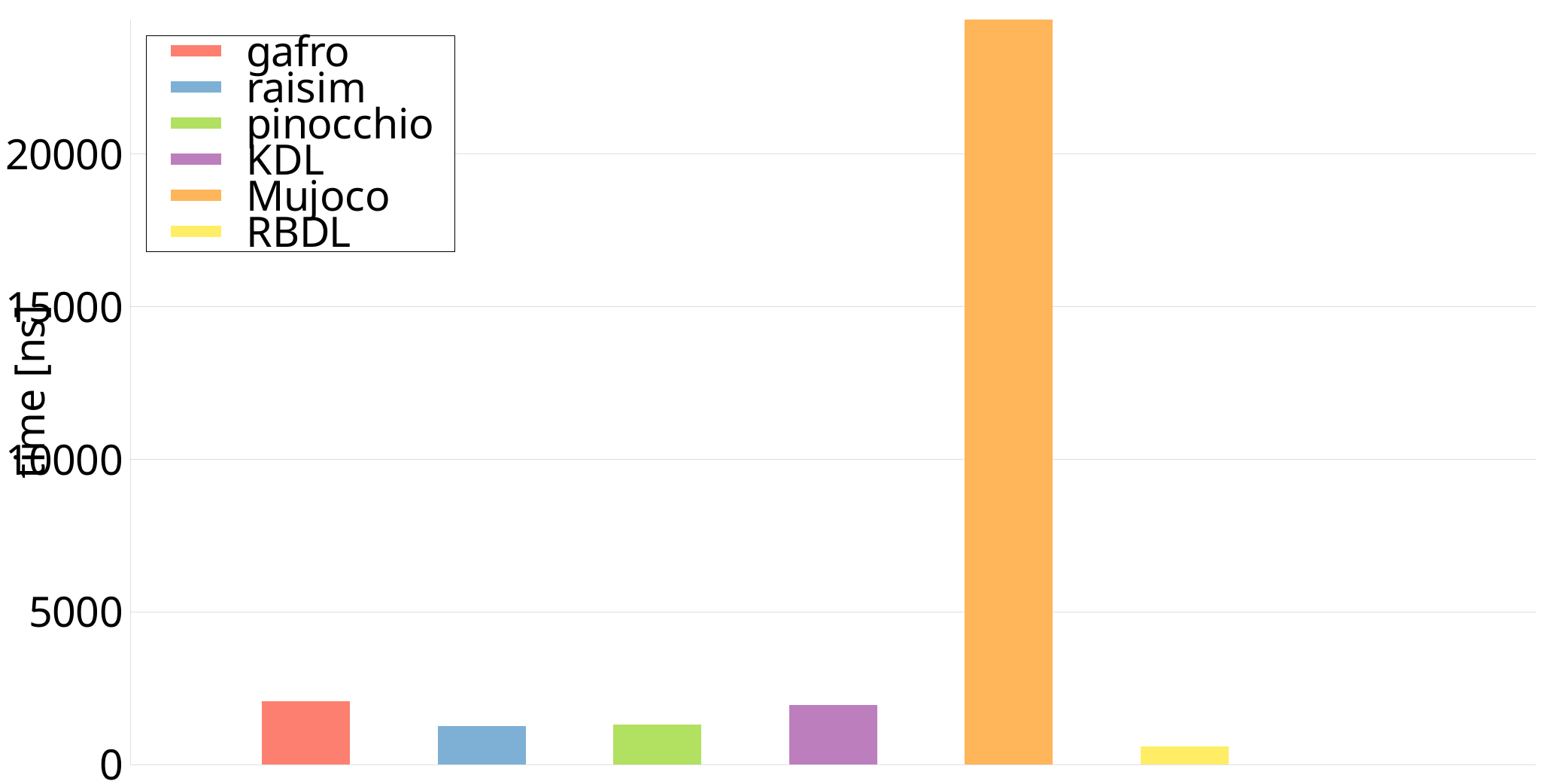}
                \caption{Inverse Dynamics.}
            \end{subfigure}
            \hfill 
            \begin{subfigure}[t]{0.475\linewidth}
                \includegraphics[width=\linewidth]{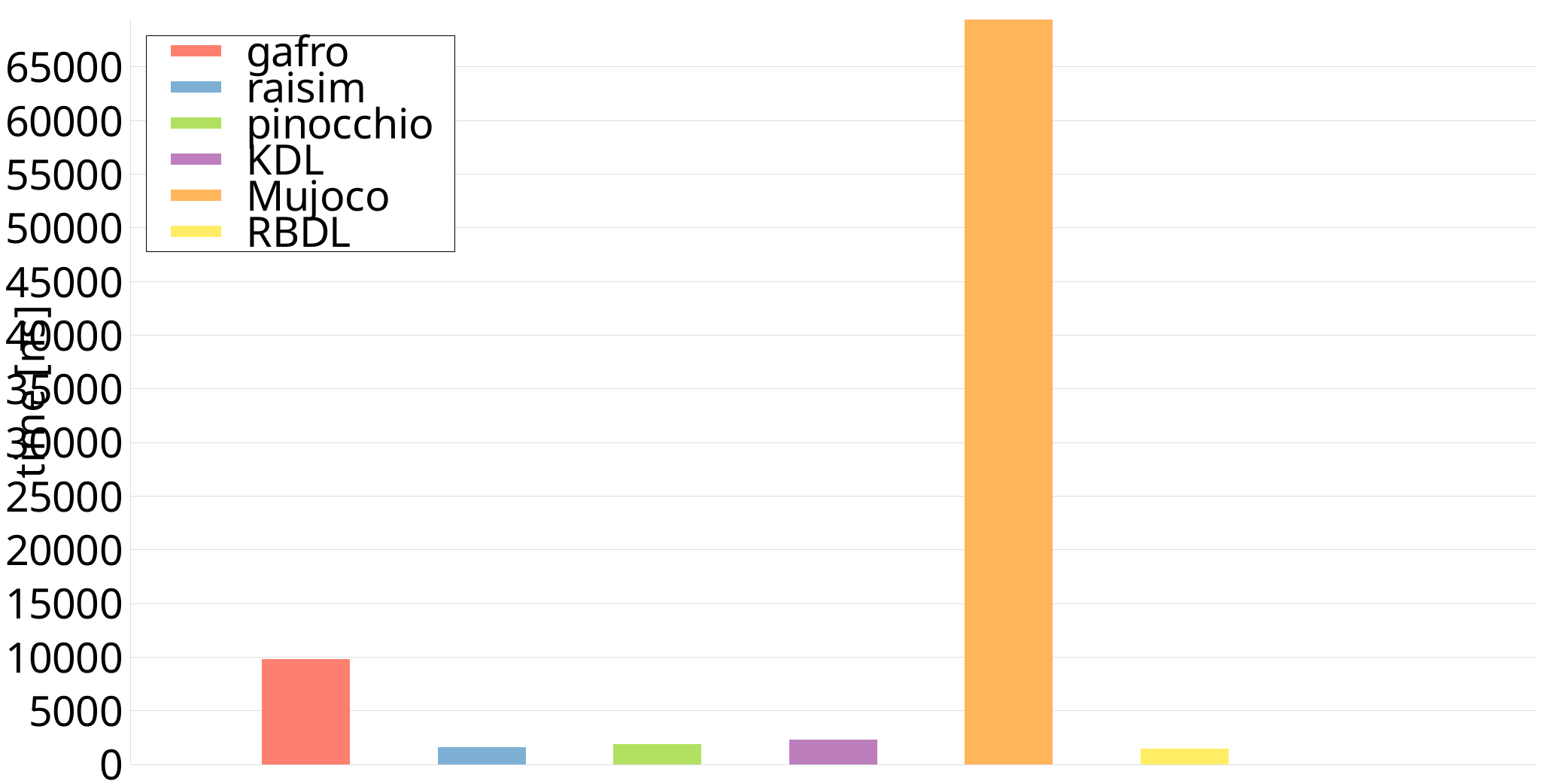}
                \caption{Forward Dynamics.}
            \end{subfigure}
            \caption{Benchmarks of robotics algorithms. The benchmark was run on an AMD Ryzen 7 4800U CPU. All libraries were compiled using \textit{gcc 13.1.1} with the compiler flags \texttt{-O3 -march=native}. The reference system is the Franka Emika Robot.}
            \label{fig:benchmarks_of_robotics_algorithms_}
        \end{figure}

        Since this library implements robot kinematics and dynamics algorithms, we are comparing and benchmarking \textit{gafro} against several libraries that are commonly used in robotics applications. This repository actually uses a git pipeline in order to continuously pull the latest changes to the libraries and update the benchmark results. The current benchmarking results on our system can be found in Figure \ref{fig:benchmarks_of_robotics_algorithms_}. As can be seen, \textit{gafro} is very competitive when it comes to the computation of the kinematics of a robotic system. The computation of the dynamics, however, especially the forward dynamics, is still slower at this point. The reason for this is an outdated implementation choice in the lower levels of the library. This issue will be addressed and fixed in a future realease of \textit{gafro}, which should make the computation of the dynamics also competitive compared to the established libraries. 

\section{APPLICATIONS AND TUTORIAL}
\label{sec:applications}
    
    In this section we provide some example applications of how the library can be used. For that purpose, we provide an accompanying repository \textit{gafro\_examples} that explains the usage of \textit{gafro}, \textit{pygafro} and \textit{gafro\_ros}. Note that in the text we are always referencing the c++ files, but the same examples can also be found in python in the corresponding folder. These examples are using the same naming scheme.

    
    \subsection{Geometric Algebra}
    \label{sub:tutorial_geometric_algebra}
        Since many potential users of \textit{gafro} are likely to be unfamiliar with the concept of geometric algebra we are providing some examples on how to do computations using this algebra. 

        
        \subsubsection{Multivector Operations}
        \label{ssub:multivector_operations}
            In a first example we are showing how to create different general multivectors and use them for algebraic computations. This example can be found in \textit{multivector.cpp}. In particular, here we show the usage of the different products in geometric algebra and how they are used with their corresponding operators that were listed in Table \ref{tab:binary_operators_}.
        
        \subsubsection{Geometric Primitives}
        \label{ssub:tutorial_geometric_primitives}
            An example on how to use the library to construct geometric primitives is shown in \textit{geometric\_primitives.cpp}. Since these geometric primitives are essentially specializations of the base \texttt{Multivector} class all operations that were presented in the example in Section \ref{ssub:multivector_operations} are also valid for the geometric primitives. Hence, we are showing the construction of the geometric primitives in different ways. First, using their algebraic definition based on the outer product and second, using their constructors. Essentially, the constructors implement the algebraic definition, and serve to simplify the usage of the library. But the example should show the equivalence of the construction. Since geometric algebra facilitates geometric reasoning, we are also providing an example of how to projections, reflections and intersections using the geometric primitives, which can be found in \textit{geometric\_primitives\_incidence.cpp}.

        
        \subsubsection{Conformal Transformations}
        \label{ssub:conformal_transformations}
            This part of the tutorial is dedicated to showing the usage of the different \texttt{Versor} classes, i.e. it shows how to apply conformal transformations to geometric primitives. There are actually two parts to this example. The first one, which is implemented in \textit{versor\_log\_exp\_map.cpp}, shows how define the different versors using the exponential mappings and recover the generator bivectors using the corresponding logarithmic mappings. In the second example, which is implemented in \textit{versors\_transforming\_geometric\_primitives.cpp}, we then show how to apply the versors to the geometric primitives. 


    
    \subsection{Robot Modeling}
    \label{sub:tutorial_robot_modeling}
        One of the targeted use cases of the \textit{gafro} library is the modeling of robotic systems. In this part of the tutorial we will show how to that in practice. 

        
        \subsubsection{Generic System}
        \label{ssub:generic_system}
            A generic system can be created using the \texttt{System} class and its \texttt{addJoint} and \texttt{addLink} functions as mentioned in Section \ref{sec:robot_modeling}. Note that in this case the parent/child relationships between the joints/links need to handled as well. An easier way is to load the system description from a yaml file, where the library will drop an error message if not all necessary relationships are defined. These two ways of creating a system are shown in the examples \textit{create\_system\_from\_code.cpp} and \textit{create\_system\_from\_file.cpp}.

        
        \subsubsection{Manipulator System}
        \label{ssub:manipulator_system}
            Manipultor systems are specialized robotic systems, which can be seen from the \texttt{Manipulator} class inheriting from the \texttt{System} class. So in order to defined a manipulator, a normal system can be created and then moved to the manipulator system with the additional information about the end-effector joint in order to create the kinematic chain. We are again providing the two examples of creating a manipulator from code or from a file in examples \textit{create\_manipulator\_from\_code.cpp} and \textit{create\_manipulator\_from\_file.cpp}, respectively.

        
        \subsubsection{Robot Kinematics/Dynamics}
        \label{ssub:robot_kinematics_dynamics}
            After creating the model of the system and in order to use the library in practical applications, the computation of the kinematics and dynamics of the systems are necessary. We show how the compute all relevant quantities using the \texttt{FrankaEmikaRobot} class in the example \textit{franka\_emika\_kinematics\_and\_dynamics.cpp}.

        
        \subsubsection{Robot Kinematics and Geometric Primitives}
        \label{ssub:robot_kinematics_and_geometric_primitives}
            In contrast to other libraries, \textit{gafro} uses geometric algebra for the modeling of robotic systems, which allows the usage of various geometric primitives at a kinematic level. In \textit{franka\_emika\_geometric\_primtives.cpp} we show an example using the \texttt{FrankaEmikaRobot} of how to move those geometric primitives to the end-effector, which will later be exploited in the formulation of the optimisation problems.

    
    \subsection{Optimisation Problems}
    \label{sub:optimization_problems}
        Many problems in important domains of robotics, such as learning and control, can be cast as optimization problems. Hence, in this section we are providing some examples on how \textit{gafro} can be used to simplify the modeling of optimisation problems using geometric algebra.

        
        \subsubsection{Inverse Kinematics}
        \label{ssub:inverse_kinematics}
            A common example in robotics is the computation of the inverse kinematics. Since we are often dealing with redundant manipulators, this becomes an optimisation problem. We are modeling this problem using the \texttt{Motor} class of \textit{gafro}, which represents poses in Euclidean space. An inverse kinematics problem in geometric algebra, formulated as an optimization problem, thus minimizes the error between two motors, which is expressed via the logarithmic map. The corresponding class that implements this cost function is called \texttt{SingleManipulatorMotorCost}. It computes the value of the cost function, as well as its gradient and hessian, to allow for first- or second-order optimisation. The example for computing the inverse kinematics using the Gauss-Newton algorithm can be found in \textit{inverse\_kinematics.cpp}. 

        
        \subsubsection{Reaching Geometric Primitives}
        \label{ssub:reaching_geometric_primitives}
            Geometric algebra extends the cost function to be uniformly applicable across the different geometric primitives. It is implemented in the class \texttt{SingleManipulatorTarget}. The template arguments \texttt{Tool} and \texttt{Target} can be different combinations of geometric primitives. Here, we cast the optimisation problem again as an inverse kinematics problem for simplicity, so we are optimizing for the joint angle configuration in which the end-effector reaches a certain geometric primitive. In previous work, however, we have shown the application of CGA to modeling manipulation tasks in an optimal control framework for model predictive control \cite{lowGeometricAlgebraOptimal2023}, which can of course be achieved using the same cost function. The different examples are listed in Table \ref{tab:example_of_optimization_problems_using_different_geometric_primitives_}.

            \begin{table}[!ht]
                \centering
                \begin{tabular}{ll}
                    \toprule
                        \textbf{Primitives}
                        & \textbf{File}
                        \\
                    \toprule
                        \texttt{Point} \& \texttt{Point}
                        & \textit{inverse\_kinematics\_point\_point.cpp}
                        \\
                    \midrule
                        \texttt{Point} \& \texttt{Line}
                        & \textit{inverse\_kinematics\_point\_line.cpp}
                        \\
                    \midrule
                        \texttt{Point} \& \texttt{PointPair}
                        & \textit{inverse\_kinematics\_point\_pointpair.cpp}
                        \\
                    \midrule
                        \texttt{Point} \& \texttt{Circle}
                        & \textit{inverse\_kinematics\_point\_circle.cpp}
                        \\
                    \midrule
                        \texttt{Point} \& \texttt{Plane}
                        & \textit{inverse\_kinematics\_point\_plane.cpp}
                        \\
                    \midrule
                        \texttt{Point} \& \texttt{Sphere}
                        & \textit{inverse\_kinematics\_point\_sphere.cpp}
                        \\
                    \midrule
                        \texttt{Line} \& \texttt{Point}
                        & \textit{inverse\_kinematics\_line\_point.cpp}
                        \\
                    \midrule
                        \texttt{Line} \& \texttt{Line}
                        & \textit{inverse\_kinematics\_line\_line.cpp}
                        \\
                    \bottomrule
                \end{tabular}
                \caption{Example of optimisation problems using different geometric primitives. These examples are located in the folder \textit{gafro\_examples/src/cpp/}. Note that with the exception of the definition of the cost function, all these files are identical.}
                \label{tab:example_of_optimization_problems_using_different_geometric_primitives_}
            \end{table}

    
    \subsection{ROS Visualization}
    \label{sub:ros_visualization}
\begin{figure}[!ht]
            \centering
            \begin{tabularx}{\linewidth}{ZZ|Z|Z}
                \includegraphics[width=0.9\linewidth]{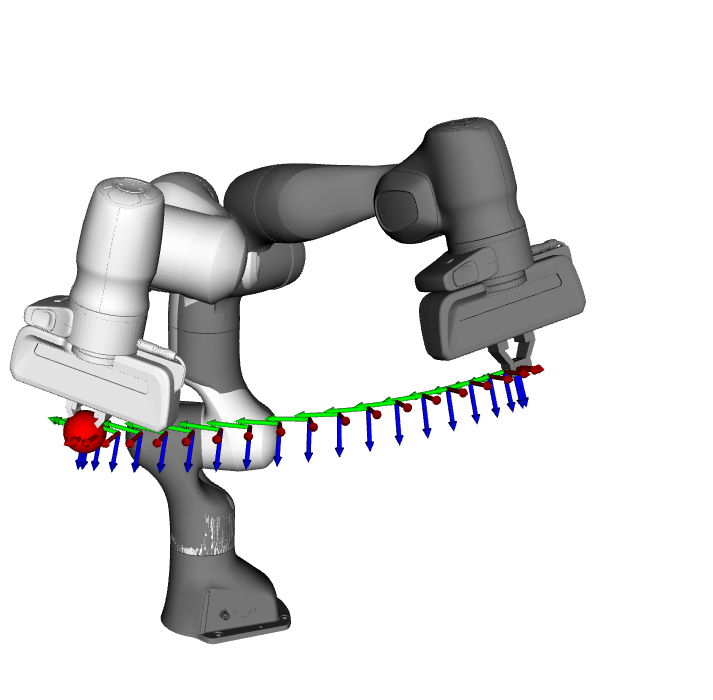}
                & \includegraphics[width=0.9\linewidth]{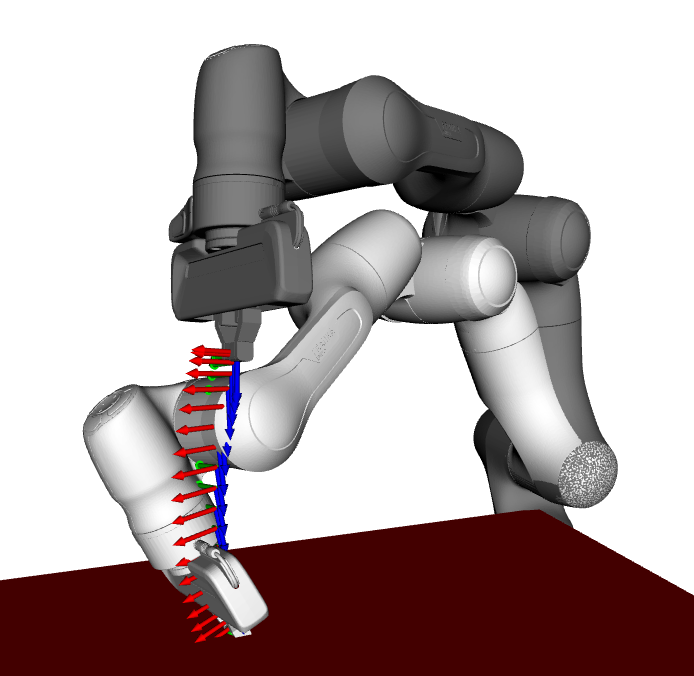}
                & \includegraphics[width=0.9\linewidth]{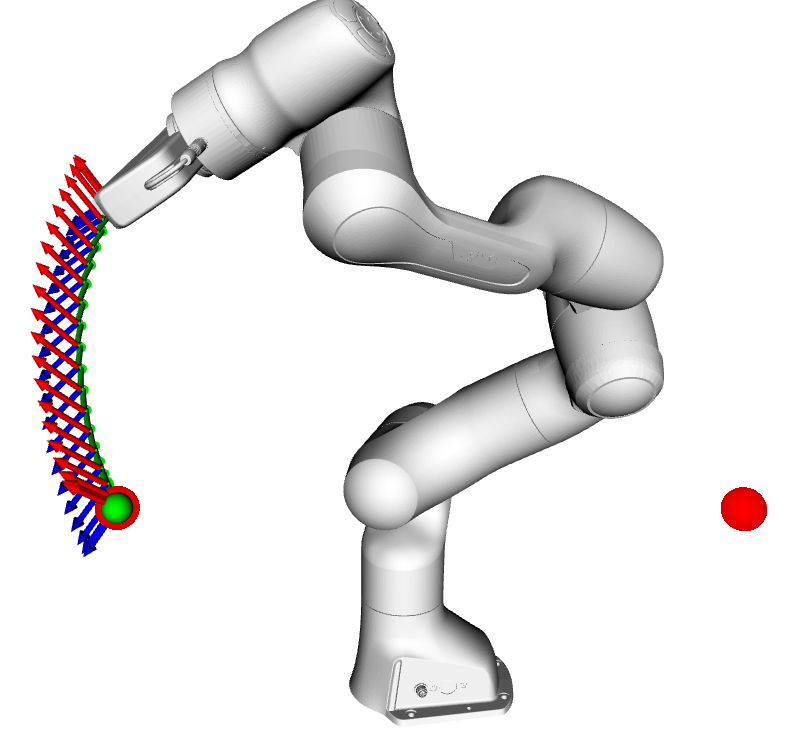}
                & \includegraphics[width=0.9\linewidth]{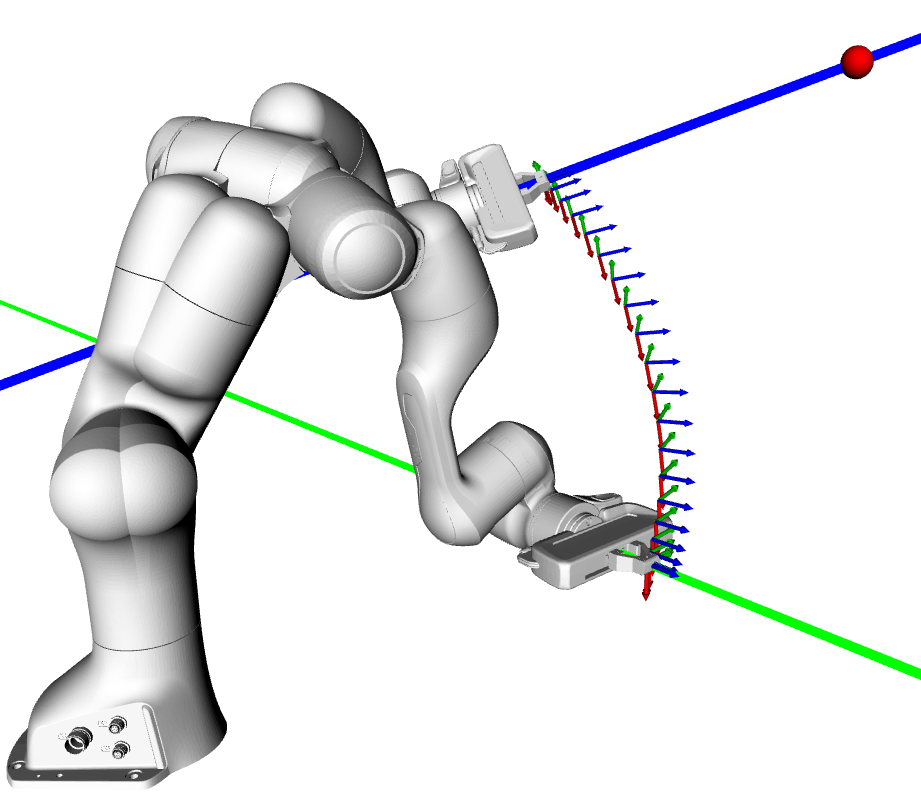}
                \\
                Point tracking
                & Plane tracking 
                & Option Point 1
                & Pointing
                \\
                \includegraphics[width=0.9\linewidth]{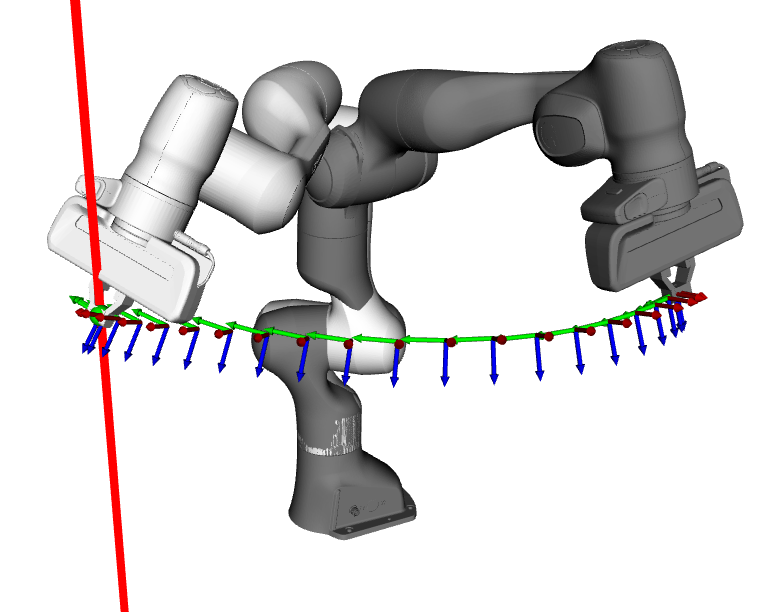}
                & \includegraphics[width=0.9\linewidth]{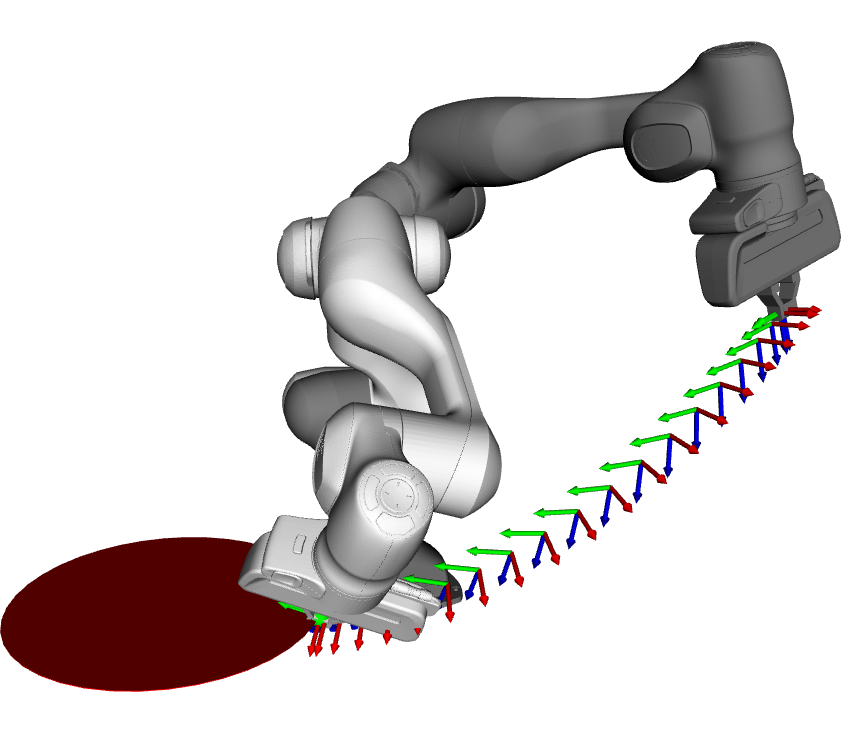}
                & \includegraphics[width=0.9\linewidth]{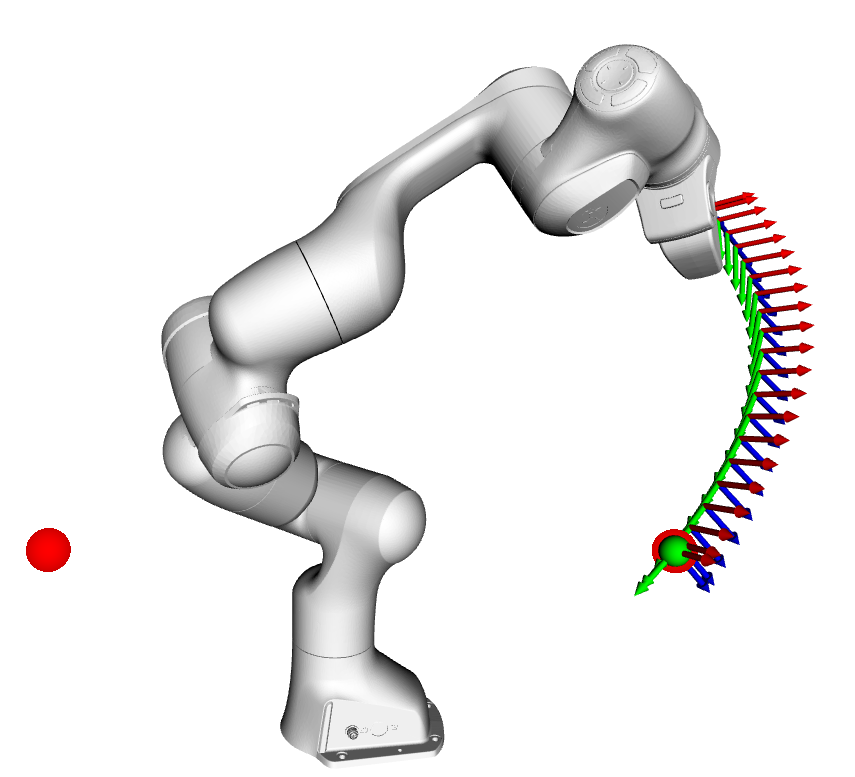}
                & \includegraphics[width=0.9\linewidth]{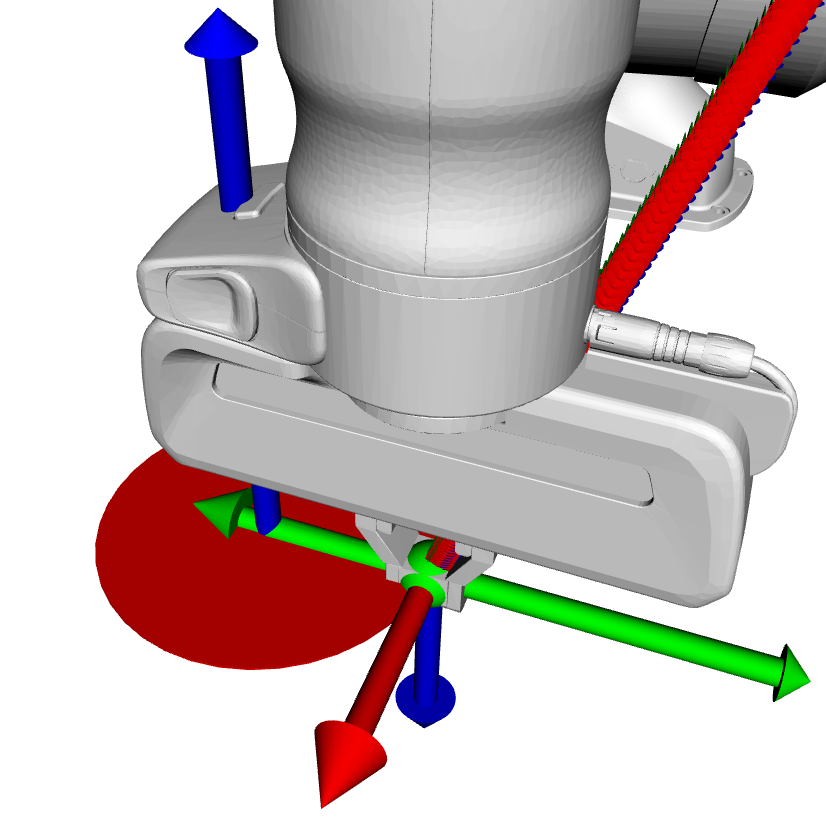}
                \\
                Line tracking
                & Circle tracking 
                & Option Point 2
                & Constraints
            \end{tabularx}
            \caption{Rviz visualizations of the Franka Emika reaching various geometric primitves.}
            \label{fig:rviz_visualizations_of_the_franka_emika_reaching_various_geometric_primitves_}
        \end{figure}

        The main purpose of the \textit{gafro\_ros} package is visualization of the geometric primitives in Rviz. Hence, we are providing an example that visualizes the Franka Emika robot reaching various geometric primitves. The example is implemented \textit{visualizing\_geometric\_primitives.cpp} and requires to be compiled in a ROS workspace. The Rviz output of the example is shown in Figure \ref{fig:rviz_visualizations_of_the_franka_emika_reaching_various_geometric_primitves_}.

\section{CONCLUSION}
\label{sec:conclusion}
    In this article we presented the implementation details as well as a tutorial for our software stack around \textit{gafro}, which is a c++ library that implements conformal geometric algebra for robotics. The software stack also includes python bindings in \textit{pygafro} as well as a ROS package in \textit{gafro\_ros}. Tutorial material and toy examples can be found in \textit{gafro\_examples}.

    While showing comparable performance for the robot modeling, geometric algebra also offers an easy and intuitive way to model various geometric relationships. Hence it provides a richer toolset than standard matrix algebra without loosing any of the existing tools. Our library \textit{gafro} provides these standard tools for robot modeling and augments them with concepts that are exclusive to geometric algebra. Providing this library that makes geometric algebra easily accessible for robotics research should allow for a wider adoption and facilitate the research on using this powerful framework for robotics. 

\printbibliography

\end{document}